\documentclass[sigconf,screen,colorlinks,authorversion=true]{acmart}
\usepackage{soul}
\usepackage{multirow}
\usepackage{color, colortbl}
\definecolor{Gray}{gray}{0.9}

\newcolumntype{L}[1]{>{\raggedright\let\newline\\\arraybackslash\hspace{0pt}}m{#1}}
\newcolumntype{C}[1]{>{\centering\let\newline\\\arraybackslash\hspace{0pt}}m{#1}}
\newcolumntype{R}[1]{>{\raggedleft\let\newline\\\arraybackslash\hspace{0pt}}m{#1}}

\copyrightyear{2023}
\acmYear{2023}
\setcopyright{acmcopyright}\acmConference[CODS-COMAD 2023]{6th Joint International Conference on Data Science \& Management of Data (10th ACM IKDD CODS and 28th COMAD)}{January 4--7, 2023}{Mumbai, India}
\acmBooktitle{6th Joint International Conference on Data Science \& Management of Data (10th ACM IKDD CODS and 28th COMAD) (CODS-COMAD 2023), January 4--7, 2023, Mumbai, India}
\acmPrice{15.00}
\acmDOI{10.1145/3570991.3570994}
\acmISBN{978-1-4503-9797-1/23/01}

\begin{document}
	
	\title{Paddy Doctor: A Visual Image Dataset for Automated Paddy Disease Classification and Benchmarking}
	
	\author{Petchiammal A}
	\affiliation{%
		\institution{Manonmaniam Sundaranar University}
		\city{Tirunelveli}
		\country{India}
	}
	\email{ampetchiammal@gmail.com}
	
	\author{Briskline Kiruba S}
	\affiliation{%
		\institution{Manonmaniam Sundaranar University}
		\city{Tirunelveli}
		\country{India}
	}
	\email{kiruba.briskline@gmail.com}
	
	\author{D. Murugan}
	%\authornote{Corresponding authors}
	\affiliation{%
		\institution{Manonmaniam Sundaranar University}
		\city{Tirunelveli}
		\country{India}
	}
	\email{dmurugan@msuniv.ac.in}
	
	\author{Pandarasamy A}
	%\authornotemark[1]
	%\authornote{Corresponding author.}
	\affiliation{%
		\country{Singapore}
	}
	\email{mkusamy@gmail.com}

	\begin{abstract}
		One of the critical biotic stress factors paddy farmers face is diseases caused by bacteria, fungi, and other organisms. These diseases affect plants' health severely and lead to significant crop loss. Most of these diseases can be identified by regularly observing the leaves and stems under expert supervision. In a country with vast agricultural regions and limited crop protection experts, manual identification of paddy diseases is challenging. Thus, to add a solution to this problem, it is necessary to automate the disease identification process and provide easily accessible decision support tools to enable effective crop protection measures. However, the lack of availability of public datasets with detailed disease information limits the practical implementation of accurate disease detection systems. This paper presents \emph{Paddy Doctor}, a visual image dataset for identifying paddy diseases. Our dataset contains 16,225 annotated paddy leaf images across 13 classes (12 diseases and normal leaf). We benchmarked the \emph{Paddy Doctor} dataset using a Convolutional Neural Network (CNN) and four transfer learning based models (VGG16,  MobileNet, Xception, and ResNet34). The experimental results showed that ResNet34 achieved the highest F1-score of 97.50\%. We release our dataset and reproducible code in the open source for community use.
	\end{abstract}
	
	
	
	
	\begin{teaserfigure}
		\includegraphics[width=\textwidth]{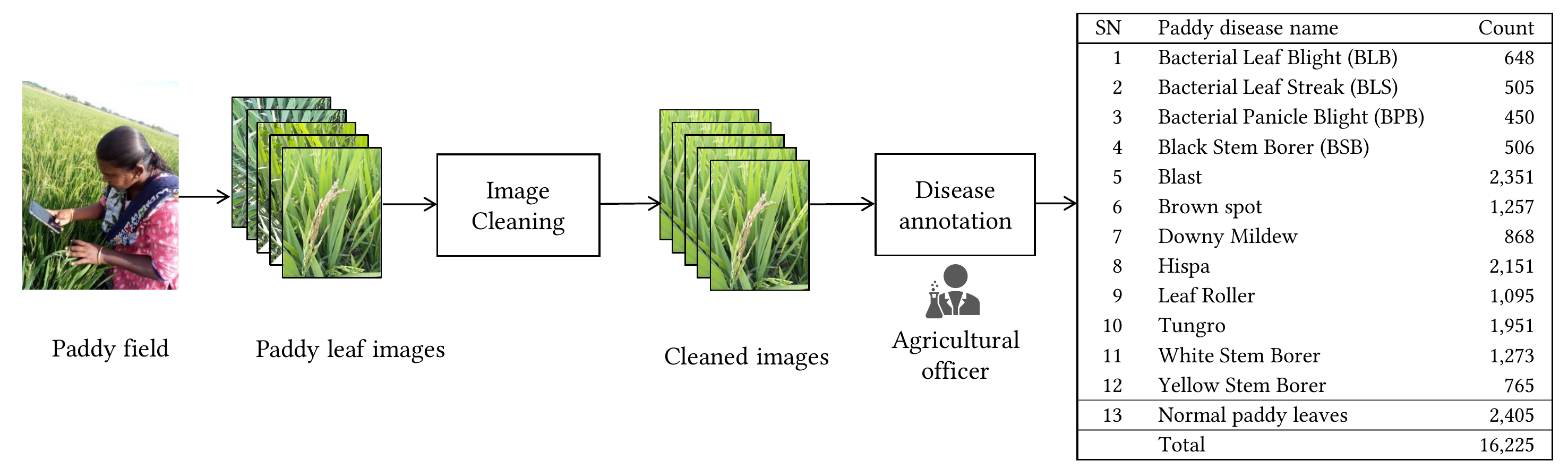}
		\caption{Data collection and annotation worflows of the \emph{Paddy Doctor} dataset (\url{https://paddydoc.github.io/}).}
		\label{fig:data-collection}
	\end{teaserfigure}
	
	\maketitle
	
	\begin{table*}[t]
		\centering
		%\begin{threeparttable}[b]
		\caption{Comparison of open-source paddy leaf disease datasets.}
		\label{Tab:dataset}
		\centering
		\begin{tabular}{L{4.0cm}R{1.8cm}R{1.2cm}R{1.2cm}L{7.8cm}}
			\toprule
			Dataset & \centering Image resolution & \centering No. of images &
			\centering No. of diseases & 
			Names of paddy diseases \\
			\midrule 
			Gujarat, India~\cite{prajapati2017detection} & 2,848 x 4,288 & 120 & 3 & Bacterial Leaf Blight (BLB), Brown spot, Leaf Smut \\ 
			%\hline
			
			Indonesia~\cite{upadhyay2022rice} & 1,440 x 1,920 & 240 & 3 &  Blast, Bacterial Leaf Blight (BLB), Tungro \\ 
			
			Malaysia~\cite{bari2021real} & 4,032 x 1,908 & 400 & 3 & Blast, Brown spot, Hispa\\     
			
			Philippines~\cite{rice552-Philippines}& 300 x 300 & 552 & 3 & Bacterial Leaf Blight (BLB), Blast, Brown spot\\ 
			
			Gujarat, India (augmented)~\cite{paddy-leaf-disease-augmented} & 3,081 x 897 & 1,294 & 3 & Bacterial Leaf Blight (BLB), Brown spot, Leaf Smut \\ 
			
			Nigeria~\cite{rice_leaf_images-Nigeria} & 256 x 256 & 3,355 & 3 & Blast, Brown spot, Hispa\\
			%\hline    
			
			%\hline
			Odisha, India~\cite{sethy2020deep} & 300 x 300 & 5,932 & 4 & Bacterial Leaf Blight (BLB), Blast, Brown spot, Tungro \\ 
			%\hline
			%\rowcolor{Gray}
			\textbf{Paddy Doctor}& \textbf{1,080 x 1,440} & \textbf{16,225} &  \textbf{12} &  \textbf{Bacterial Leaf Blight (BLB), Bacterial Leaf Streak (BLS), Bacterial Panicle Blight (BPB), Black Stem Borer, Blast, Brown spot, Downy Mildew, Hispa, Leaf Roller, Tungro, White Stem Borer, Yellow Stem Borer}\\ 
			%  \rowcolor{Gray}
			%  \textbf{Paddy Doctor}&\textbf{480 x 640}&\textbf{17,324}&\textbf{13}&\textbf{BLB, BLS, Blast, Brown spot, Hispa, Leaf roller, Mildew on Rice Leaf, Rice Black Stem Borer, Rice Stem Borer Egg, Rice Yellow Stem Borer, Tungro}\\
			%\hline
			\bottomrule
			%\multicolumn{4}{c}{BLB: Bacterial Leaf Blight; BLS: Bacterial Leaf Streak; BPB: Bacterial Panicle Blight.}
		\end{tabular}
		% \begin{tablenotes}
		%     \small
		%     \item []BLB: Bacterial Leaf Blight; BLS: Bacterial Leaf Streak; BPB: Bacterial Panicle Blight. 
		%     \item []
		% \end{tablenotes}    
		%\end{threeparttable}
	\end{table*}

	\section{Introduction}
	\label{sec:introduction}
	Agriculture is one of the most important industries contributing to the majority of the national income in several countries. In India alone, 70\% of the rural population relies on agriculture~\cite{foaindia}. Paddy is a ubiquitous crop in most Asian countries, and India is the world's second-largest producer of paddy. Paddy cultivation is a complex process affected by many diseases and pests. The early identification of these paddy diseases is a daunting task for agriculturists as well as for agriculture experts~\cite{shrivastava2019rice}. Traditionally, farmers employ manual techniques based on their experience and visual inspection to identify the paddy diseases, but this is highly inefficient, time-consuming, and error-prone~\cite{leelavathy2021prediction}. At times, even experienced farmers and agriculture experts might fail to identify the crop diseases accurately due to the large variety of identical disease symptoms.
	Moreover, farmers apply a large quantity of fertilizer or pesticide without identifying the exact reason for disease manifestation, monitoring the depth of the disease, and measuring the micro-nutrient deficiency. Pesticides are well known for affecting both plants and the soil. It is increasingly important to automate the process of detection of the paddy disease at the earlier stage to reduce pesticide usage and subsequently minimize the loss in the yield~\cite{haridasan2023deep}.

	With the advent of Information and Communication Technology (ICT), many researchers have proposed automated disease identification methods by leveraging computer vision techniques~\cite{ferentinos2018deep}. While the traditional methods usually involve manual feature engineering, the recent deep learning-based approaches automatically extract and analyze image features and improve performance. Convolutional neural networks are one of the widely used techniques. Moreover, the variations of convolutional neural network architecture such as DenseNet~\cite{verma2021leaf}, AlexNet~\cite{zakzouk2021rice}, and EfficientNet~\cite{atila2021plant} have enabled the machines to understand critical patterns from the diseased part of the leaf, delivering even better performances than human analysis in many classification problems. Despite all these efforts, the lack of availability of labeled data from real paddy fields hinders the proliferation of these techniques into practical use.

	This paper presents \emph{Paddy Doctor}, a large-scale annotated dataset for automated paddy disease identification. The paddy leaf images were collected from real paddy fields using high-resolution smartphone cameras. The collected images were carefully cleaned and annotated with the help of an agricultural officer. The final dataset contains 16,225 leaf images across 13 classes (12 distinct diseases and healthy leaves). Furthermore, we benchmark our \emph{Paddy Doctor} dataset using five advanced state-of-the-art deep-learning models and compare their performance. The models are Deep Convolutional Neural Network (DCNN) and four transfer learning-based models such as VGG16, MobileNet, Xception, and ResNet34\footnote{\url{https://keras.io/api/applications/}}. Our experimental results revealed that ResNet34 achieved the highest F1-score of 97.50\%. We release the \emph{Paddy Doctor} dataset and reproducible code in the open source\footnote{\url{https://paddydoc.github.io/}}.
	
	The rest of the paper is organized as follows. In Section~\ref{sec:related}, we review the related works. In Section~\ref{sec:dataset}, we describe our \textit{Paddy Doctor} dataset in detail. In Section~\ref{sec:benchmarking}, we present our benchmarking study and results, followed by conclusions in Section~\ref{sec:conclusion}.

	\section{Related Work}
	\label{sec:related}
	A few public datasets are available to experiment with and develop automated paddy disease classification systems. Table~\ref{Tab:dataset} compares the image resolution, number of images, number of diseases, and list of paddy diseases present in the existing public datasets. In~\cite{prajapati2017detection}, authors have prepared a database of 120 paddy leaf images (40 samples each for three diseases) in Gandhinagar, Gujarat, India. An augmented version of the same dataset containing 1,294 images is available in ~\cite{paddy-leaf-disease-augmented}. Similarly, authors from Indonesia~\cite{upadhyay2022rice}, Philippines~\cite{rice552-Philippines}, and Nigeria~\cite{rice_leaf_images-Nigeria} have also created a public dataset of 240, 552, and 3,355 images, respectively, across three disease classes. It is to be noted that each image in these datasets contains a close-up view of a single paddy leaf, showing disease symptoms on white background, captured in a controlled environment using high-resolution professional cameras. Unlike this, the researchers from Malaysia~\cite{bari2021real} have created a dataset of 400 images, across three disease classes, by capturing the paddy leaf images from real paddy fields. Similarly, in ~\cite{sethy2020deep}, a large public dataset with 5,932 images from Odisha, India, is presented. They used a high-resolution professional camera for data collection and then extracted the patches (300x300) of the diseased portion from the original large images to create the final annotated dataset. 
	
	Inline with the ongoing efforts towards dataset creation, quite a few research groups have also developed machine learning and deep learning based techniques to detect and classify rice diseases~\cite{sagarika2020paddy,wang2021rice,swathika2021disease,patil2022rice}. In~\cite{sagarika2020paddy}, a CNN model was used to classify three types of rice diseases. The proposed CNN model has achieved an accuracy of 94.12\%. In \cite{wang2021rice}, the authors proposed an Attention-based Depthwise Separable Neural Network - Bayesian Optimization (ADSNN-BO) model that has achieved an accuracy of 94.65\%. In \cite{bharathi2020paddy}, AlexNet model was used for rice disease identification and achieved an accuracy of 96.5\%. In \cite{patil2022rice}, CNN and Multilayer Perceptron (MLP) models were proposed to achieve 81.03\% and 91.25\% accuracy respectively.
	
	Due to lack of availability, most of the prior work used relatively smaller datasets with fewer paddy diseases (See Table~\ref{Tab:dataset}). In contrast, we present a large paddy leaf disease dataset containing 16,225 annotated images with 13 classes (12 diseases and normal leaf) and also benchmark the performance of several off-the-shelf deep-learning models.

	\begin{figure}[t!]
		\centering
		\includegraphics[scale=0.72]{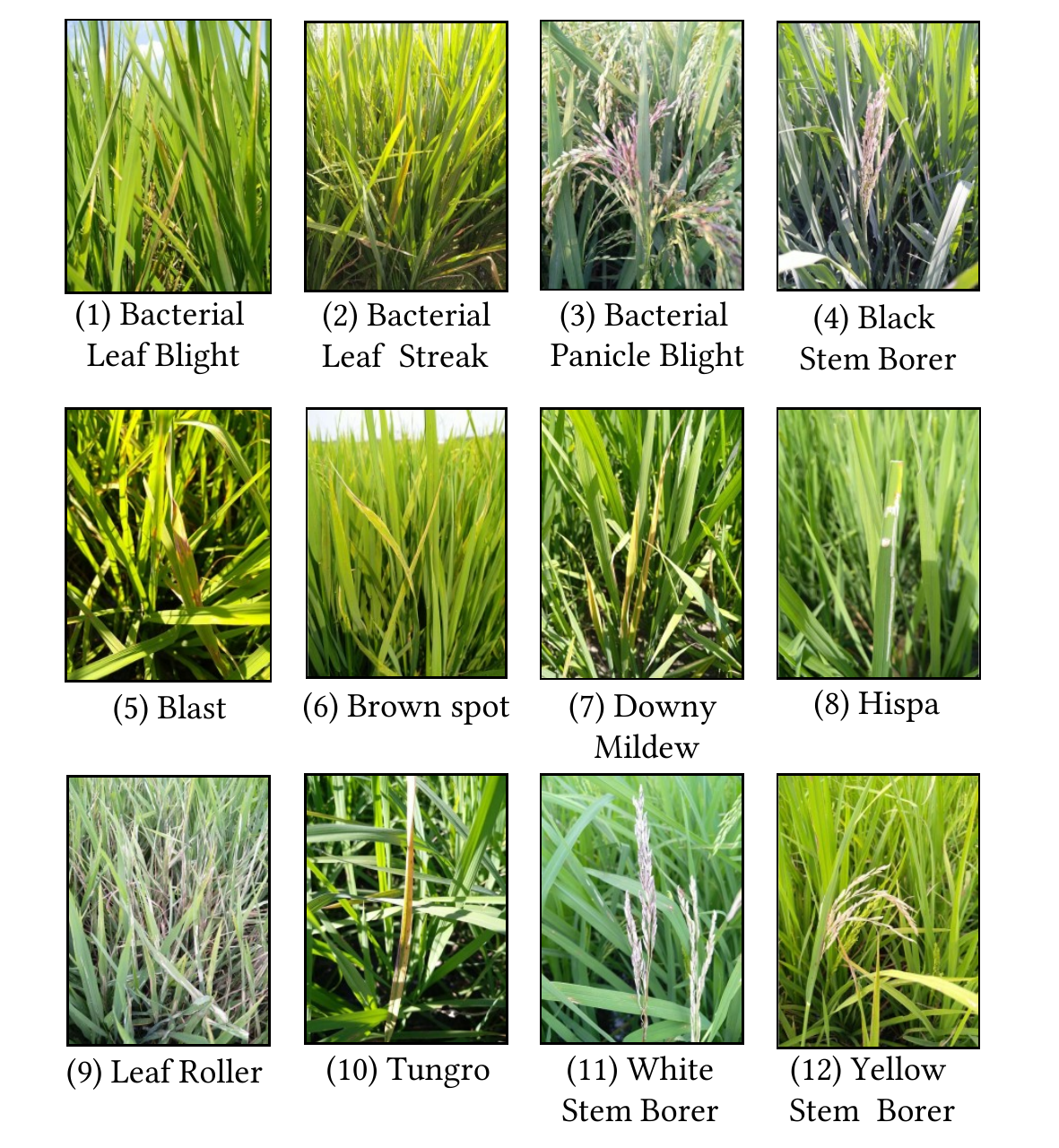}
		\caption{Sample disease images from \emph{Paddy Doctor} dataset.}
		\label{fig:sample}
	\end{figure}

	\begin{figure*}[t!]
		\centering
		\includegraphics[scale=0.5]{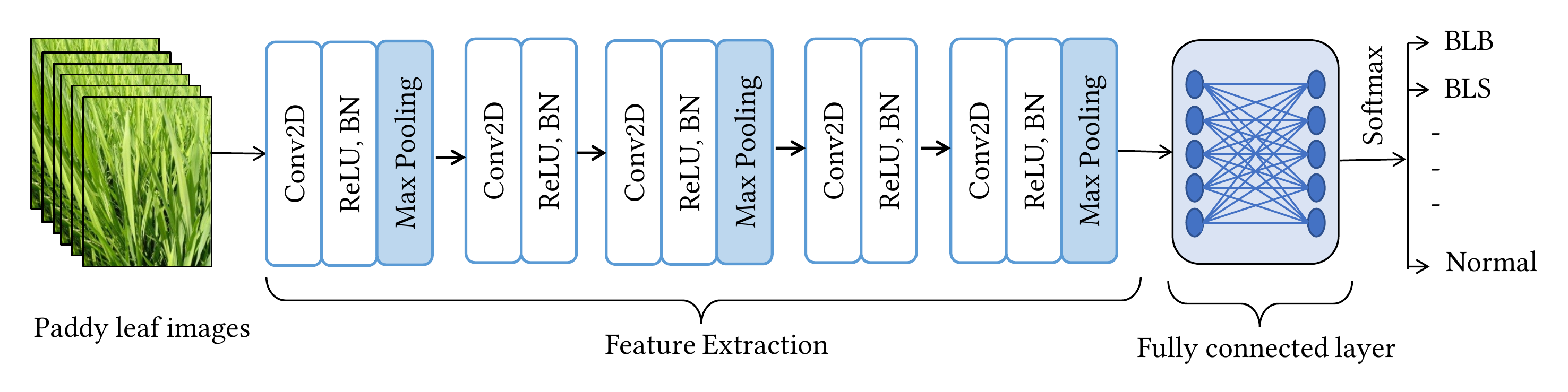}
		\caption{A six-layer deep CNN model for paddy disease classification.}
		\label{fig:cnn}
	\end{figure*}
	
	\begin{figure*}[t!]
		\centering
		\includegraphics[scale=0.5]{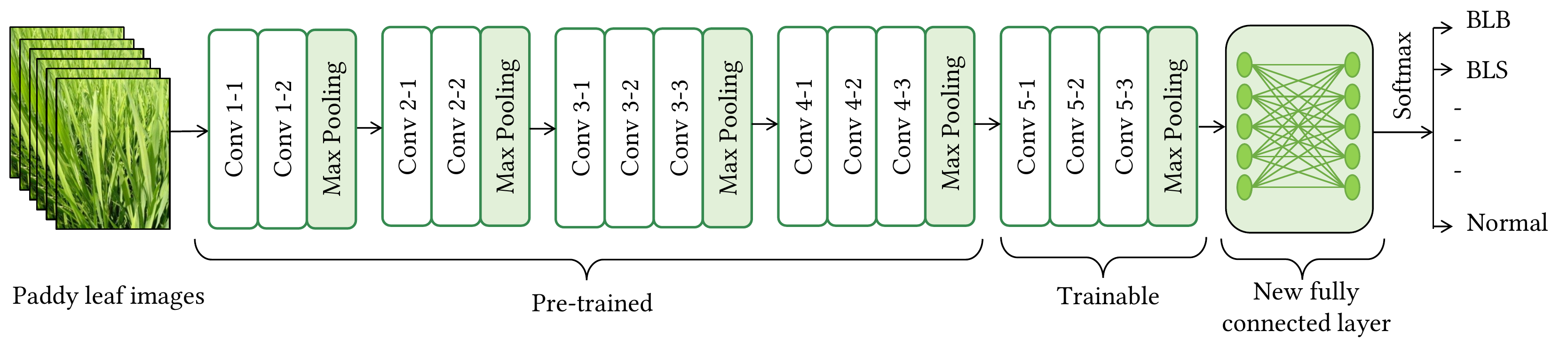}
		\caption{Fine-tuning of VGG16 model for paddy disease classification.}
		\label{fig:tl}
	\end{figure*}
	
	\section{Paddy Doctor Dataset}
	\label{sec:dataset}
	The data collection and annotation process of the \emph{Paddy Doctor} dataset is shown in Figure~\ref{fig:data-collection}. We collected RGB images of paddy leaves from real paddy fields in a village near the Tirunelveli district of Tamilnadu, India. The data collection happened from February to April 2021, when the age of the paddy crop was between 40 to 80 days. We used the CAT S62 Pro smartphone with a built-in camera to capture high-resolution RGB images. Our initial dataset contained approximately 30,000 images in JPEG format with a pixel resolution of 1,080 (width) by 1,440 (height). Next, we carefully examined each sample and removed the inferior and duplicate images. After image cleaning, we are left with 16,225 images in our dataset.
	
	Next, we manually annotated each image, with the help of an agricultural officer, based on the presence of disease symptoms and assigned a diseased class label. After annotation, the final dataset had 13 classes, corresponding to 12 diseases and healthy leaves. The annotated paddy diseases are as follows - Bacterial Leaf Blight (BLB), Bacterial Leaf Streak (BLS), Bacterial Panicle Blight (BPB), Black Stem Borer (BSB), Blast, Brown spot, Downy Mildew, Hispa, Leaf Roller, Tungro, White Stem Borer, Yellow Stem Borer, and Normal leaf (See Figure~\ref{fig:data-collection}). 
	
	Figure~\ref{fig:sample} shows the sample images of the leaves having 12 distinct diseases. In addition to the RGB images, we manually collected additional metadata for each leaf image, such as the variety and age of the paddy crop when these images were logged. The entire dataset development process spanned approximately 500 man-hours.
	
	\section{Benchmarking}
	\label{sec:benchmarking}
	We benchmark our {Paddy Doctor} dataset using five contemporary deep-learning models and compare their performance in classifying paddy disease images. The experimented models include a deep CNN model and transfer learning with four pre-trained models: VGG16,  MobileNet, Xception, and ResNet34. The details of these models are presented below.
	
	\subsection{Deep Convolutional Neural Network}
	The architecture of the CNN model is shown in Figure~\ref{fig:cnn}. It consists of five 2D convolutional layers and a final dense layer. The first convolutional layer is filtered with 32 kernels of size 3 × 3. Then, a 3×3 max-pooling layer is added after the first convolutional layer. The next convolutional layer contains 64 convolution kernels of size 3×3. We have used a batch normalization layer to automatically standardize the inputs in a model and improve the accuracy and stability of neural networks. Similarly, the subsequent three layers use filters of sizes 64, 128, and 128, respectively. The last layer is composed of a max-pooling layer. Two dense connectivity strategies improve the usage efficiency of feature maps, enhancing the diagnostic performance for paddy leaf diseases and a 13-way Softmax layer.
	
	\subsection{Transfer Learning}
	In addition to the CNN model, we also apply four existing deep learning models to our dataset and evaluate their performance. Though the pre-training models can be used as feature extractors and predictors, we fine-tuned them to perform better. As shown in Figure~\ref{fig:tl}, the fine-tuning approach involves keeping most of the existing pre-trained convolutional layers but training only the last few layers and a custom fully connected layer. The selected pre-trained models are VGG16, MobileNet, Xception, and ResNet34. We initialized the weights of these models using ImageNet\footnote{\url{https://image-net.org/}}. Therefore, the training phase of these models involved assigning new weights to the last few layers and the final fully connected layer. The code repository of the \emph{Paddy Doctor} contains more details about their configurations.
	
	\begin{table}
		\centering
		\caption{Comparison of classification performance of five deep learning models on our \textit{Paddy Doctor} dataset. The ResNet34 model achieved the best performance with an F1-score of 97.50\%.}
		\label{tab:metrics}
		\begin{tabular}{rlrrrr}
			\toprule
			\multirow{2}{*}{S.No.} &
			\multirow{2}{*}{Model} & \multicolumn{4}{c}{Metrics (\%)} \\
			& & Accuracy & Precision & Recall & F1-score \\
			\midrule
			1 & DCNN & 88.84 & 89.22 & 88.84 & 88.81 \\
			2 & MobileNet & 92.42 & 92.63 & 92.42 & 92.39 \\
			3 & VGG16 & 93.19 & 93.49 & 93.19 & 93.20 \\
			4 & Xception & 96.58 & 96.61 & 96.58 & 96.57 \\
			\rowcolor{Gray}
			5 & ResNet34 & 97.50 & 97.52 & 97.50 & 97.50 \\
			\bottomrule
		\end{tabular}
	\end{table}
	
	\subsection{Experimental Results}
	We implemented the five deep learning models in Python framework using Keras and TensorFlow libraries and conducted all the experiments on the Google Collab environment with GPU. We split the entire \textit{Paddy Doctor} dataset into two sets: training and testing. The training set had 12,980 (80\%), and the test set had 3,245 (20\%) images out of the total 16,225 images. In addition, we extracted a validation set consisting of 2956 (20\%) images from the training set itself. Therefore, the final training set had 10,384 images. We also used image data augmentation during model training by applying different image transformation techniques. The operations include rotation (5\textdegree), shear intensity (0.2\textdegree), zoom (0.2), width and height shift (5\%), and horizontal flip. Moreover, all images were resized into 256x256 pixels and normalized. All models used a learning rate of 0.001, 100 epochs, and a batch size of 32 during training. 
	
	Table~\ref{tab:metrics} compares the classification performance of five deep learning models using four evaluation metrics: accuracy, precision, recall, and F1-score. We observed that the ResNet34 model achieved the highest F1-score of 97.50\%. This is followed by Xception (96.57\%), VGG16 (93.20\%), and MobileNet (92.39\%). Comparatively, the DCNN model achieved the lowest F1-score of 88.81\%. These results demonstrate the usability of our \emph{Paddy Doctor} dataset for automated paddy disease classification tasks. Additionally, we plan to evaluate other pre-trained models leveraging different transfer learning strategies in the future.
	
	\section{Conclusion}
	\label{sec:conclusion}
	Manual identification of paddy diseases is a challenging task for farmers. Hence, there is an increasing need to develop automated solutions that can scale to many diseases and plants. The lack of availability of public datasets with annotated disease names was a major bottleneck to benchmarking the recent deep learning-based models and wider adoption of the solutions. In this paper, we presented the \emph{Paddy Doctor} dataset for automated paddy disease detection. It contains 16,225 annotated paddy leaf images across 13 classes (12 diseases and normal leaf). The presented dataset was benchmarked using five deep learning-based models and we compared their performance across each other. The results demonstrate that ResNet34 achieved a superior accuracy of 97.5\% followed by 96.58\% with Xception based model. Finally, plans are underway to expand our \emph{Paddy Doctor} dataset by collecting fine-grained data, such as infrared and hyper-spectral images, about paddy diseases and pests and benchmark them using additional deep learning models.
	
	\bibliographystyle{ACM-Reference-Format}
	\bibliography{references}
	
\end{document}